\documentclass{ecai} 

\usepackage{latexsym}
\usepackage{amssymb}
\usepackage{amsmath}
\usepackage{hyperref}
\usepackage{amsthm}
\usepackage{booktabs}
\usepackage{enumitem}
\usepackage{graphicx}
\usepackage{color}
\usepackage{subcaption}

\usepackage[utf8]{inputenc}

\usepackage{microtype}

\usepackage{inconsolata}

\begin{document}


\begin{frontmatter}




\title{Controlling Summarization Length Through EOS Token Weighting}


\author[A]{\fnms{Emmanouil}~\snm{Stergiadis}\footnote{Equal contribution.}}
\author[A]{\fnms{Zeno}~\snm{Belligoli}\thanks{Corresponding Author. Email: zeno.belligoli@booking.com}\footnotemark}
\author[A]{\fnms{Eran}~\snm{Fainman}}
\author[A]{\fnms{Ilya}~\snm{Gusev}} 

\address[A]{Booking.com}


\begin{abstract}
Controlling the length of generated text can be crucial in various text-generation tasks, including summarization. Existing methods often require complex model alterations, limiting compatibility with pre-trained models. We address these limitations by developing a simple approach for controlling the length of automatic text summaries by increasing the importance of correctly predicting the \(EOS\) token in the cross-entropy loss computation. The proposed methodology is agnostic to architecture and decoding algorithms and orthogonal to other inference-time techniques to control the generation length. We tested it with encoder-decoder and modern GPT-style LLMs, and show that this method can control generation length, often without affecting the quality of the summary.
\end{abstract}

\end{frontmatter}


\section{Introduction}

Text summarization is the task of condensing essential information from a long text into a shorter one. Extractive text summarization methods create summaries by taking the most representative sentences from the original text, whereas abstractive text summarization focuses on generating completely new text \cite{abstractive}. This task finds applications in various domains, such as news \cite{hermann}, scientific papers \cite{luhn}, conversations \cite{samsum}, and review \cite{reviews} summarization.

Summarization tasks tend to be accompanied by various constraints, often dictated by the requirements of an application or product. Examples of these constraints are capping the maximum length of the generated text, using specific keywords in the summary, following a specific format or style \cite{controllable}. 

Furthermore, despite the rise of large language models like ChatGPT or GPT-4 \cite{openai2023gpt4}, we speculate (and confirm in Section \ref{results}) that simpler models can offer comparable summarization quality at a lower cost, making research in this field still relevant.

In this work, we focus on controlling length in abstractive text summarization. This problem is motivated by the need to meet interface requirements, such as element sizes in mobile applications. In this context, summaries need to be of a desired character length to fit into the page to optimize user experience.

To address this problem, we introduce a simple method of controlling summary length which involves weighting the end-of-sentence (\( EOS \)) token more than other tokens at training time. Intuitively, this allows the model to focus on correctly predicting when to stop the generation, thus inducing it to respect the summary length distribution in its training data. We conduct experiments on two model families and multiple decoding strategies to show the portability of the proposed approach across architectures and its complementarity to other inference-time length controlling techniques. 


\section{Previous work}

Methods for controlling the length of generated text can be categorized into two groups: learning-based and decoding-based approaches. Although learning-based methods involve alterations to the training architecture or loss function, decoding-based methods operate during the inference phase.

Decoding-based techniques often involve preventing the model from producing the \( EOS \) token by assigning it a probability of negative infinity and truncating the text once the desired token count is achieved \cite{rush2015neural}, or by incorporating a length penalty into the beam-search decoding algorithm \cite{murray2018correcting}.

On the other hand, learning-based methods adapt the attention mechanism to be more sensitive to length \cite{yu2021lenatten, liu2022length} or train specialized embeddings that factor in the desired length of the generated text \cite{kikuchi2016controlling, fan2017controllable, liu2018controlling, takase2019positional}. In addition, \citet{makino2019global} designed a modification of the objective function that increases the effectiveness of embedding-based methods, thus showing that the modifications of the training architecture and of the objective function are complementary to each other. Many of these techniques, however, entail intricate implementation steps and necessitate training new models from scratch, making them less feasible for integration with pretrained models. 

Notable exceptions to this constraint are the work of \citet{miculicich2023summarization}, who fine-tuned a pre-trained model with reversed positional encodings and showed competitive results both in terms of summary quality and length, and that of \citet{chan-etal-2021-controllable} and \citet{jie2023prompt} who used a Markov decision process and reinforcement learning, respectively, to control the generation length. 

In line with this research trajectory, our method can be applied to train a new model from scratch as well as to fine-tune pretrained models. We refrain from altering the underlying architecture, and instead adopt a straightforward modification of the objective function which enhances our ability to govern generation length.


\section{Methodology}\label{sec:methodology}

The intuition behind our method lies in the special importance of the \( EOS \) token during training. We note that the cross-entropy loss calculated on that particular token is the only loss component directly teaching the model to respect the summary length distribution in its training data\footnote{This is not the case when the summary consists of a single sentence terminated by a period. In that case, the period token acts as an extra, albeit somewhat weaker, \( EOS \) token.}. During the computation of the loss, the signal from that particular token gets diluted by the averaging operation among all other generated tokens, which depending on the dataset can range in number from a few dozens to a few hundreds.

We therefore hypothesize that simply boosting the weight of that loss component will help the model follow the training length distribution more closely, without significantly affecting overall performance. To be precise, our work aims at enforcing an upper bound for the generation length, which is why we are only interested in penalising false negatives when predicting \( EOS \) token (the loss component when the ground truth \textbf{is} \( EOS \)). The exact weight to be applied is a hyper-parameter on which we run an ablation study.

In formal terms, we start from the original form of the cross entropy loss calculated over the sequence:
\begin{equation}
L_{1} =  - \frac{1}{N}\sum_{n=1}^{N} \log \frac{e^{x_{n}^{y_n}}}{\sum_{v=1}^{|V|} e^{x_{n}^{v}}}
\end{equation}

where \( V \) is the vocabulary, \( N \) is the sequence length, \( y_{n} \) the ground truth token at time-step \( n \in (1, N) \) and \( x_{n}^{v} \) is the logit for token \( v \in V \) at time-step \( n \). We then add a weighting term to derive:

\begin{equation} L_{2} =  - \frac{R}{N}\sum_{n=1}^{N} w_{y_{n}}\log \frac{e^{x_{n}^{y_n}}}{\sum_{v=1}^{|V|} e^{x_{n}^{v}}}\end{equation} 

where

\[
    w_{y_{n}}= 
\begin{cases}
    W,& \text{if } y_{n} = [EOS]\\
    1,              & \text{otherwise}
\end{cases}
\]
\newline

Because this weighting marginally impacts the norm of the loss and therefore its gradient, we apply a re-scaling factor

\begin{equation} R = \frac{N}{N + W - 1} \end{equation} 

to make sure the norm of weight updates are not impacted in expectation. This effectively changes the mean pooling of loss components $1 / N$ in the original loss to $1 / (N + W - 1)$ which represents a loss computed over $N + W - 1$ components: the $N - 1$ non $EOS$ ones and the $EOS$ one counted $W$ times.

The weight of the \( EOS \) token \(W\) is a hyper-parameter that controls the balance between semantics and length: when \(W = 1\), \(L_2\)  goes back to treating \( EOS \) just as another token (\(L_1 = L_2\)); as \(W \to \infty\) the loss assigns higher importance to not missing the \( EOS \) token, thus making its predicted sequences increasingly short (potentially at the expense of quality).

\section{Experiments}\label{sec:experiments}

We devised two variants of the proposed methodology based on the availability of datasets with different characteristics. For this, we create subsets of \textbf{CNN/Daily Mail} \cite{hermann2015teaching, seeetal2017get} and \textbf{XL-sum} \cite{hasan2021xl} with the desired characteristics. We consider only the English part of XL-sum and remove all summaries consisting of just one sentence (see Section \ref{sec:methodology} for the explanation).

\subsection{Fixed-Length Approach}
This variant requires datasets with summaries that respect the desired length constraint. Hence, we randomly select samples with a summary length of 250 and 175 characters or less for CNN/Dailymail and XL-sum, respectively. The training, validation, and test sets of both datasets consist of 10k, 500, and 500 samples.

\subsection{Dynamic-Length Approach}
This variant circumvents the need for manually curating datasets with a specific summary length by pre-pending the instruction '\texttt{Summarize with up to \{K\} characters the following text:}' to each sample in the dataset, where \texttt{K} is the number of characters in the reference summary. This would induce the model to "learn to count" the number of characters at inference time, thus being able to generate summaries of any desired length. For CNN/Dailymail we gather 100k, 500, and 500 samples for the train, validation and test set. For XL-sum we gather 20k, 500, and 500 samples for the train, validation and test set. For every sample, we prepend the prompt above and, for simplicity, round \texttt{K} up to the closest number in the range between 50 and 800 with a stride of 50 for CNN/Dailymail, and in the range between 25 and 400 with a stride of 25 for XL-sum.

\subsection{Base Models and Hyperparameters}\label{sec:models_and_hp}
For both variants, we fine-tune the pre-trained T5-base \cite{2020t5} and Llama-2 7B \cite{touvron2023llama} models with values of the \( EOS \) token weight of 1 (baseline) and 10. We compare two decoding strategies: greedy decoding, and beam search with 5 beams and length penalty values of -1, 0 and 1. 

We use the Hugging face\footnote{https://huggingface.co/} framework to fine-tune our models. We use the AdamW optimizer \cite{loshchilov2017decoupled} with a learning rate of 5e-5 and weight decay of 0.01. We reduce the learning rate using a cosine scheduler for Llama-2 7B and a linear one for T5-base. We train using an effective batch size of 2 on all models for a maximum of 10,000 steps and pick the best checkpoint in terms of validation loss. We use a maximum source length of 4096 tokens and target max length of 512 tokens. For Llama-2 7B, instead of fine-tuning the full network, we use qLoRA adapters \cite{dettmers2024qlora} for every linear layer with a $r$ of 16, an $\alpha$ of 16, and dropout of 0.05. 

As baselines, we use gpt-3.5-turbo and gpt-4o by OpenAI\footnote{https://openai.com/} with default generation parameters and the following prompt template prepended and appended to the input text: \verb|"Summarize with up to {K} characters:|".\footnote{We tested different positions for the prompt and found that both prepending and appending it to the input text yields the best results.}

\subsection{Metrics}
As metrics, we report (a) \textbf{ROUGE-N} \cite{lin2004rouge}: a relevance score for text generation tasks which relies on the intersection of N-grams between the reference and prediction. Since we observed a strong correlation between ROUGE metrics, we report only ROUGE-2 in the main paper and the full suite in Appendix~\ref{sec:full}; (b) \textbf{BERTScore} \cite{zhang2019bertscore}: a semantic similarity score calculated using contextual embeddings from a pre-trained BERT model, in our case the 40th layer of Deberta-xlarge-mnli \cite{deberta} as it correlated the best with human judgement in the \href{https://docs.google.com/spreadsheets/d/1RKOVpselB98Nnh_EOC4A2BYn8_201tmPODpNWu4w7xI/edit#gid=0}{WMT-16 benchmark}; (c) \textbf{Percent of too long summaries}: the percent of generated summaries that exceed the number of character limitation. This is our primary metric.

\section{Results}\label{results}

\begin{figure}[!h]
\setlength{\belowcaptionskip}{10pt}
\centering
\begin{subfigure}[b]{\linewidth}
    \centering
    \includegraphics[width=\linewidth]{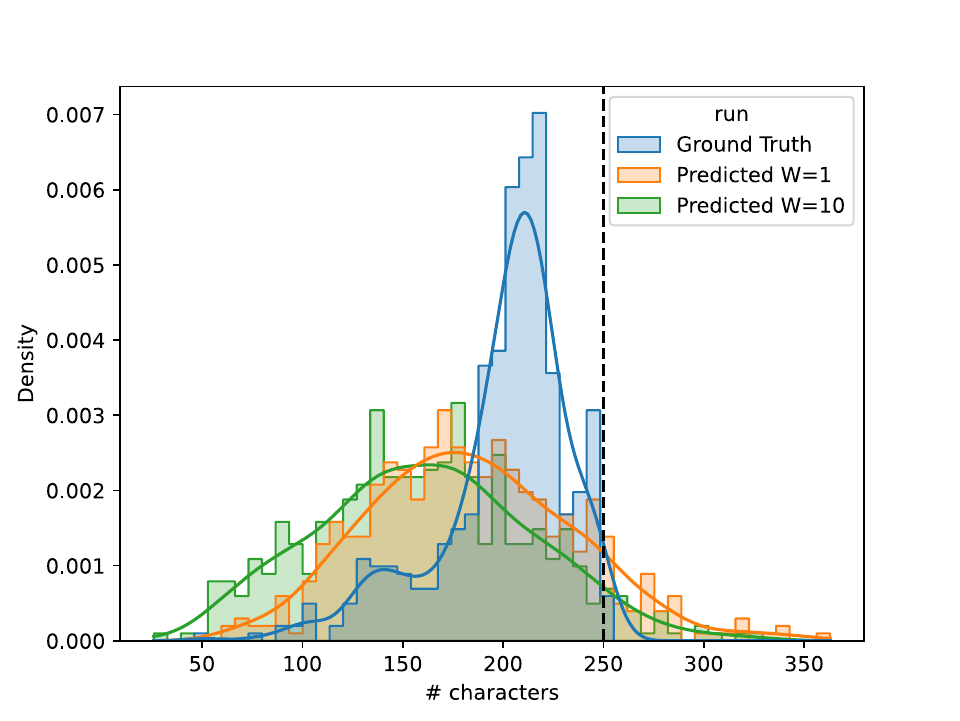}
    \caption{T5-base}
    \label{fig:t5base_histplot}
\end{subfigure}
\begin{subfigure}[b]{\linewidth}
    \centering
    \includegraphics[width=\linewidth]{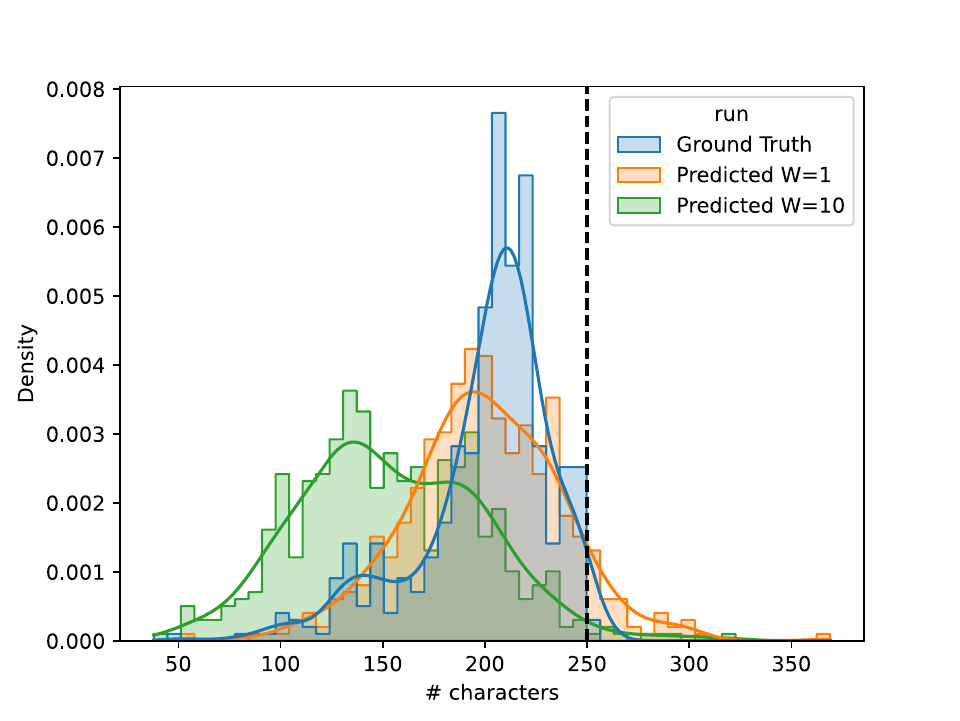}
    \caption{Llama-2 7B}
    \label{fig:llama2_histplot}
\end{subfigure}
\caption{Length distributions of predicted test summaries with different EOS weights for CNN/Dailymail.}
\label{fig:combined_figures}
\end{figure}

Table~\ref{tab:results_cnn_ablation1} shows how metrics differ across several $W$ settings. As expected, higher values result in better length control by shifting the distribution of generated length to the left as shown in Figure~\ref{fig:combined_figures}. However we note there are diminishing returns after a certain value of \(W\) which in our setting lies somewhere between 10 and 100. This is also why we fixed $W=10$ for all subsequent experiments.
\begin{table}[!h]
\small
\begin{center}
    \begin{tabular}{|c|c|c|c|}
        \hline
        $w$ & Rouge-2 & BertScore & \% of long \\
        \hline
        \multicolumn{4}{|c|}{\textbf{T5-base}} \\
        \hline
        1 & 14.7 & 26.1 & 9.8 \\
        10 & 14.5 & 26.0 & 5.4 \\
        100 & 14.3 & 25.6 & 8.6 \\
        1000 & 14.3 & 26.1 & 2.2 \\
        \hline
        \multicolumn{4}{|c|}{\textbf{Llama-2-7B}} \\
        \hline
        1 & 17.3 & 34.6 & 7.6 \\
        10 & 17.0 & 33.2 & 1.8 \\
        100 & 15.4 & 30.0 & 0.4 \\
        1000 & 12.2 & 26.5 & 0.0 \\
        \hline
    \end{tabular}
    \vspace{10pt}
    \caption{Results for CNN/Daily Mail, Fixed Length approach (250 characters), different EOS weights and greedy decoding.}
    \label{tab:results_cnn_ablation1}
\end{center}
\end{table}

The results for the Fixed-length approach are shown in Tables~\ref{tab:results_cnn} and \ref{tab:results_xlsum}. We observe that the proposed method always controls length better than the baseline, across architectures and decoding strategies. For T5-base, our method does not show significant degradation of summary quality across all settings, both in terms of Rouge-2 and BertScore. For Llama-2-7B, on the other hand, there seems to be often a trade off between summary quality and length control. 

We want to ensure that the length decreasing mechanism learned using our method is not trivial, i.e. similar to a simple truncation baseline. We include that baseline (truncate the text at exactly 250 characters for the CNN/Daily Mail dataset) and measure the percentage of generated summaries that do not end with a punctuation mark. This is a proxy for unnaturally truncated text, an undesirable effect. Table \ref{tab:results_cnn_truncation} shows that, unlike the naive baseline, our method does not unnaturally cut-off sentences.

\begin{table}[!h]
\small
\begin{center}
    \begin{tabular}{|c|c|}
    \hline
        method & \% cut-off \\
    \hline
        \multicolumn{2}{|c|}{\textbf{T5-base}} \\
    \hline
        $w=1$           &     7.0     \\ \hline
        $w=10$          &     8.4     \\  \hline
        truncation      &    16.2     \\ \hline
        \multicolumn{2}{|c|}{\textbf{Llama-2-7B}} \\
    \hline
        $w=1$            &    4.6     \\ \hline
        $w=10$           &    4.6     \\ \hline
        truncation       &    11.6    \\ \hline
    \end{tabular}
    \vspace{10pt}
    \caption{Effect of truncation and EOS weighting on cut-off sentences and length on the CNN/Daily Mail dataset. Greedy decoding.}
    \label{tab:results_cnn_truncation}
\end{center}
\end{table}

Tables~\ref{tab:results_cnn} and \ref{tab:results_xlsum} show that the positive effects of our method are consistent across decoding strategies and, in particular, are present even when beam search with length penalty\footnote{The \textit{lp} parameter is actually a length reward as implemented in HuggingFace, i.e. positive values penalise short, rather than long generations} is used, proving that our method is orthogonal to inference-time length control techniques. We also note that gpt-3.5-turbo and gpt-4o failed to adhere to the specified length constraints provided via prompts. Both models show inferior performance compared to our fine-tuned Llama-2 7B and T5-base across all metrics.
\begin{table}[!h]
\small
\begin{center}
    \resizebox{0.482\textwidth}{!}{%
        \begin{tabular}{|l|c|c|c|c|c|c|}
            \hline
                & \multicolumn{2}{|c|}{Rouge-2} & \multicolumn{2}{|c|}{BertScore} & \multicolumn{2}{|c|}{\% of too long} \\
                & $w$=$1$ & $w$=$10$ & $w$=$1$ & $w$=$10$ & $w$=$1$ & $w$=$10$\\
         \hline
         \multicolumn{7}{|c|}{\textbf{T5-base}} \\
         \hline
         Greedy & \textbf{14.7} & 14.5 & \textbf{26.1} & 26.0 & 9.8 & \textbf{5.4}  \\
         $\text{Beam}_{-1}$ & \textbf{15.7} & 15.2 & \textbf{27.4} & 26.9 &  7.0 & \textbf{2.8}  \\
         $\text{Beam}_{0}$ & 15.6 & \textbf{15.7} & \textbf{27.4} & 26.9 & 10.2 & \textbf{4.2}  \\
         $\text{Beam}_{+1}$ & 15.4 & \textbf{15.7} & 25.5 & \textbf{26.1} & 57.4 & \textbf{37.0}  \\
         \hline
         \multicolumn{7}{|c|}{\textbf{Llama-2-7B}} \\
         \hline
         Greedy & \textbf{17.3} & 17.0 & \textbf{34.6} & 33.2 & 7.6 & \textbf{1.8}  \\
         $\text{Beam}_{-1}$ & \textbf{16.4} & 15.6 & \textbf{30.6} & 28.6 & 1.4 & \textbf{0.0}  \\
         $\text{Beam}_{0}$ & \textbf{16.7} & 15.3 & \textbf{30.8} & 28.3 & 1.0  & \textbf{0.0}  \\
         $\text{Beam}_{+1}$ & \textbf{16.7} & 15.5 & \textbf{30.8} & 28.6 & 2.0 & \textbf{0.0}  \\
         \hline
         \multicolumn{7}{|c|}{OpenAI} \\
         \hline
         gpt-3.5-turbo & \multicolumn{2}{|c|}{12.3} & \multicolumn{2}{|c|}{28.1} & \multicolumn{2}{|c|}{36.2} \\
         gpt-4o & \multicolumn{2}{|c|}{12.9} & \multicolumn{2}{|c|}{29.1} & \multicolumn{2}{|c|}{76.8} \\
         \hline
        \end{tabular}
        }
    \vspace{10pt}
    \caption{Results for modified CNN/Daily Mail, Fixed Length approach (250 characters). The subscripts in Beam denote the value of the length penalty parameter.}
    \label{tab:results_cnn}
\end{center}
\end{table}
\begin{table}[!h]
\small
\begin{center}
    \resizebox{0.482\textwidth}{!}{%
        \begin{tabular}{|l|c|c|c|c|c|c|}
            \hline
                & \multicolumn{2}{|c|}{Rouge-2} & \multicolumn{2}{|c|}{BertScore} & \multicolumn{2}{|c|}{\% of too long} \\
                & $w$=$1$ & $w$=$10$ & $w$=$1$ & $w$=$10$ & $w$=$1$ & $w$=$10$\\
         \hline
         \multicolumn{7}{|c|}{\textbf{T5-base}} \\
         \hline
         Greedy & 11.9 & \textbf{12.0} & 33.2 & \textbf{33.9} & 2.0 & \textbf{0.6}  \\
         $\text{Beam}_{-1}$ & 12.7 & \textbf{12.8} & 36.2 & 36.2 & 0.0 & 0.0  \\
         $\text{Beam}_{0}$ & \textbf{13.0} & 12.9 & \textbf{36.3} & 36.2 & 0.0 & 0.0  \\
         $\text{Beam}_{+1}$ & 12.9 & \textbf{13.0} & 33.1 & \textbf{34.4} & 8.2 & \textbf{4.8}   \\
         \hline
         \multicolumn{7}{|c|}{\textbf{Llama-2-7B}} \\
         \hline
         Greedy & \textbf{17.5} & 16.9 & \textbf{43.9} & 43.6 & 1.0 & \textbf{0.8}  \\
         $\text{Beam}_{-1}$ & \textbf{18.3} & 18.2 & 44.0 & 44.0 & 0.0 & 0.0  \\
         $\text{Beam}_{0}$ & 18.2 & 18.2 & \textbf{44.0} & 43.9 & 0.0 & 0.0  \\
         $\text{Beam}_{+1}$ & 18.3 & 18.3 & \textbf{44.0} & 43.9  & \textbf{0.0} & 0.2  \\
         \hline
         \multicolumn{7}{|c|}{OpenAI} \\
         \hline
         gpt-3.5-turbo & \multicolumn{2}{|c|}{4.6} & \multicolumn{2}{|c|}{24.5} & \multicolumn{2}{|c|}{11.4} \\
         gpt-4o & \multicolumn{2}{|c|}{4.3} & \multicolumn{2}{|c|}{24.6} & \multicolumn{2}{|c|}{27.4} \\
         \hline
        \end{tabular}
    }
    \vspace{10pt}
    \caption{Results for XLsum-multi-sentence, Fixed Length approach (175 characters).}
    \label{tab:results_xlsum}
\end{center}
\end{table}
 
\begin{figure}[!h]
\setlength{\belowcaptionskip}{10pt}
\centering
\begin{subfigure}[b]{\linewidth}
    \centering
    \includegraphics[width=\linewidth]{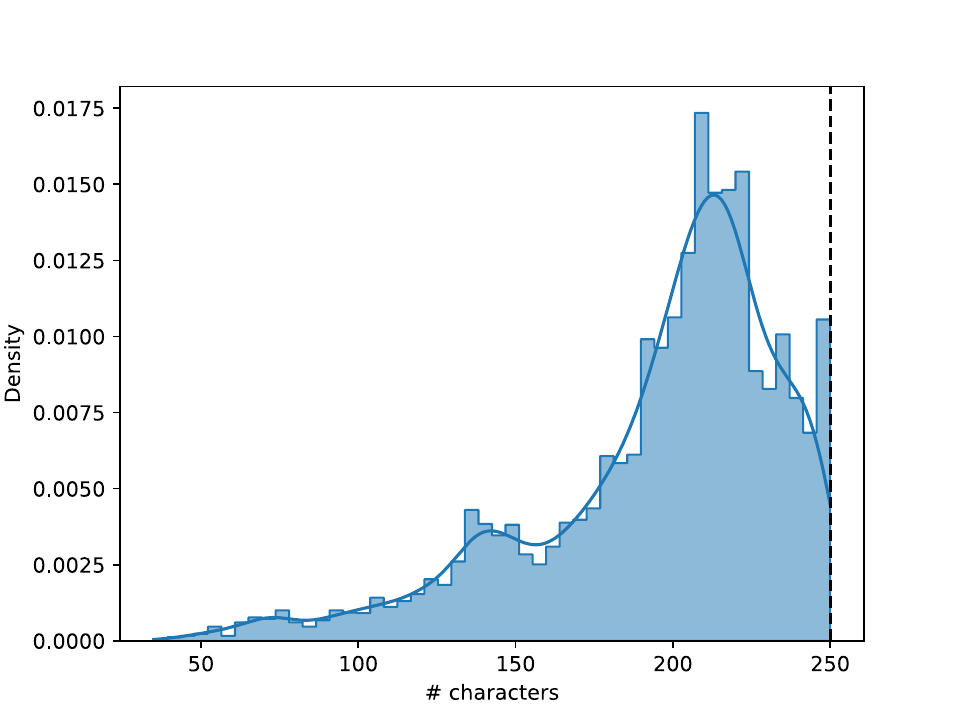}
    \caption{CNN/Dailymail training set distribution.}
    \label{fig:cnn_train_histplot}
\end{subfigure}
\begin{subfigure}[b]{\linewidth}
    \centering
    \includegraphics[width=\linewidth]{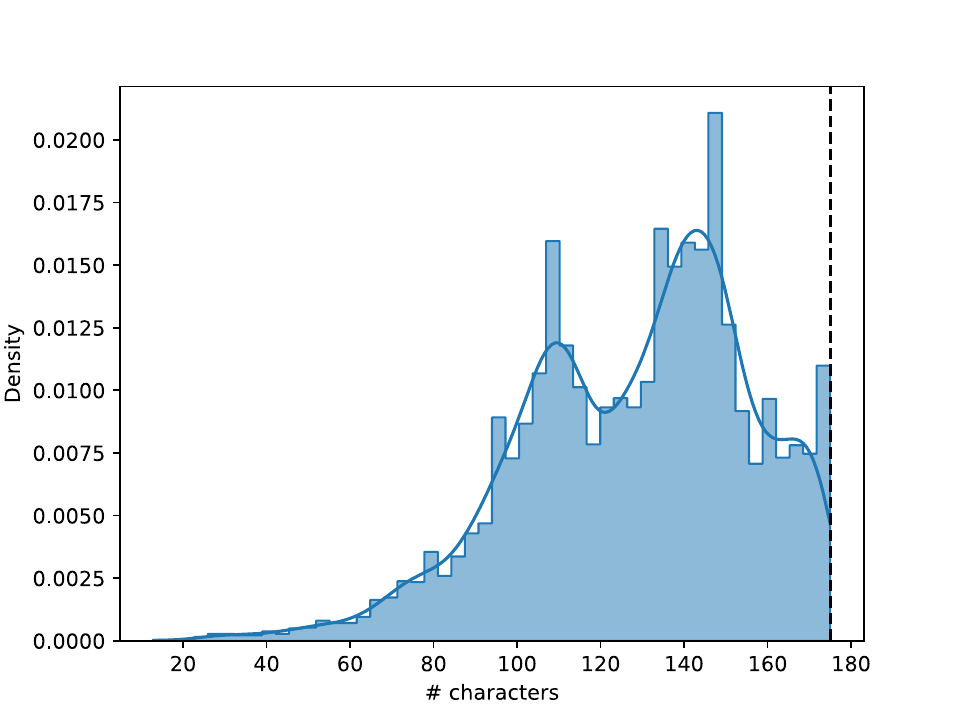}
    \caption{XLsum-multi-sentence training set distribution.}
    \label{fig:xlsum_train_histplot}
\end{subfigure}
\caption{Length distributions of summaries in training sets.}
\label{fig:combined_figures_train}
\end{figure}

Tables~\ref{tab:results_cnn_dynamic} and \ref{tab:results_xlsum_dynamic} show the results for the  Dynamic Length variant. For CNN/Dailymail, we observe the proposed method significantly improves length control over the baseline. In addition, it also improves summary quality for T5-base but not for Llama2 7B. On XL-sum, our approach achieves comparable summary quality to the baseline for T5-base, and consistently better summary quality for Llama-2 7B. However, it fails to improve on length control with respect to the baseline. This is unexpected. We speculate this may be due to the summary length distribution of XL-sum being heavily right-skewed with a bimodal distribution as shown in Figure~\ref{fig:combined_figures_train} (shown for the Fixed length dataset). The presence of a consistent number of summaries far below the length threshold nudges the model towards producing short text. This is in contrast to the distribution of CNN/Dailymail, for which most of the mass is concentrated close to the threshold.

\begin{table}[!h]
\small
\begin{center}
    \resizebox{0.482\textwidth}{!}{%
        \begin{tabular}{|l|c|c|c|c|c|c|}
        \hline
            & \multicolumn{2}{|c|}{Rouge-2} & \multicolumn{2}{|c|}{BertScore} & \multicolumn{2}{|c|}{\% of too long} \\
            & $w$=$1$ & $w$=$10$ & $w$=$1$ & $w$=$10$ & $w$=$1$ & $w$=$10$\\
         \hline
         \multicolumn{7}{|c|}{\textbf{T5-base}} \\
         \hline
         Greedy & 15.6 & \textbf{16.3} & 25.5 & \textbf{26.8} & 37.2 & \textbf{18.6}  \\
          $\text{Beam}_{-1}$ & 16.6 & 16.6 & 27.8 & \textbf{27.9} & 34.0 & \textbf{13.4}  \\
          $\text{Beam}_{0}$ & 16.6 & \textbf{16.8} & 27.8 & \textbf{27.9} & 36.4 & \textbf{20.0}  \\
          $\text{Beam}_{+1}$ & 15.9 & \textbf{16.9} & 24.6 & \textbf{27.4} & 83.4 & \textbf{61.2}  \\
         \hline
         \multicolumn{7}{|c|}{\textbf{Llama-2-7B}} \\
         \hline
         Greedy & \textbf{18.8} & 18.7 & \textbf{35.9} & 35.5 & 21.0 & \textbf{12.4}  \\
          $\text{Beam}_{-1}$ & \textbf{18.6} & 18.1 & \textbf{34.1} & 32.8 & 14.8 & \textbf{8.4} \\
          $\text{Beam}_{0}$ & \textbf{18.7} & 18.1 & \textbf{34.1} & 32.9 & 17.0 & \textbf{9.6} \\
          $\text{Beam}_{+1}$ & \textbf{18.7} & 18.1 & \textbf{34.2} & 32.8 & 19.6 & \textbf{11.0} \\
          \hline
         \multicolumn{7}{|c|}{OpenAI} \\
         \hline
         gpt-3.5-turbo & \multicolumn{2}{|c|}{12.1} & \multicolumn{2}{|c|}{28.8} & \multicolumn{2}{|c|}{18.0} \\
         gpt-4o & \multicolumn{2}{|c|}{13.4} & \multicolumn{2}{|c|}{30.9} & \multicolumn{2}{|c|}{53.0} \\
         \hline
        \end{tabular}
        }
    \vspace{10pt}
    \caption{Results for CNN/Daily Mail, Dynamic Length approach with \texttt{K} \texttt{in} \texttt{range(start=50, stop=800, step=50)}.}
    \label{tab:results_cnn_dynamic}
\end{center}
\end{table}
\begin{table}[!h]
\small
\begin{center}
    \resizebox{0.482\textwidth}{!}{%
        \begin{tabular}{|l|c|c|c|c|c|c|}
        \hline
            & \multicolumn{2}{|c|}{Rouge-2} & \multicolumn{2}{|c|}{BertScore} & \multicolumn{2}{|c|}{\% of too long} \\
            & $w$=$1$ & $w$=$10$ & $w$=$1$ & $w$=$10$ & $w$=$1$ & $w$=$10$\\
         \hline
         \multicolumn{7}{|c|}{\textbf{T5-base}} \\
         \hline
         Greedy & 10.9 & \textbf{11.2} & 32.4 & \textbf{32.6} & \textbf{10.4} & 11.4  \\
          $\text{Beam}_{-1}$ & \textbf{12.5} & 12.2 & \textbf{34.4} & 34.1 & \textbf{7.2} & 7.6  \\
          $\text{Beam}_{0}$ &  \textbf{12.7} & 12.3 & \textbf{34.4} & 34.2 & \textbf{7.2} & 8.0  \\
          $\text{Beam}_{+1}$ &  \textbf{12.6} & 12.3 & 31.6 & \textbf{31.7} & 24.4 & \textbf{23.8}  \\
         \hline
         \multicolumn{7}{|c|}{\textbf{Llama-2-7B}} \\
         \hline
         Greedy & \textbf{16.5} & 15.7 & 41.6 & 41.6 & \textbf{7.2} & 9.2  \\
          $\text{Beam}_{-1}$ & 17.1 & \textbf{17.5} & 42.0 & \textbf{42.2} & \textbf{3.4} & 3.6  \\
          $\text{Beam}_{0}$ & 17.1 & \textbf{17.5} & 41.8 & \textbf{42.2} & \textbf{3.4} & 3.8  \\
          $\text{Beam}_{+1}$ & 17.0 & \textbf{17.5} & 41.8 & \textbf{42.2} & \textbf{3.6} & 4.0  \\
         \hline
         \multicolumn{7}{|c|}{OpenAI} \\
         \hline
         gpt-3.5-turbo & \multicolumn{2}{|c|}{3.8} & \multicolumn{2}{|c|}{22.5} & \multicolumn{2}{|c|}{29.0} \\
         gpt-4o & \multicolumn{2}{|c|}{3.3} & \multicolumn{2}{|c|}{22.5} & \multicolumn{2}{|c|}{33.0} \\
         \hline
        \end{tabular}
        }
    \vspace{10pt}
    \caption{Results for XLsum-multi-sentence, Dynamic Length approach with \texttt{K} \texttt{in} \texttt{range(start=25, stop=400, step=25)}.}
    \label{tab:results_xlsum_dynamic}
\end{center}
\end{table}

\section{Conclusions}

This paper introduced a simple and effective method for controlling text summarization length: increasing the weight of the EOS token in the training loss function. Our experiments across diverse models (T5-base, Llama-2 7B) and decoding strategies demonstrated that this technique significantly improves adherence to length constraints, often without a substantial loss in summary quality as measured by ROUGE-2 and BERTScore.

The proposed EOS weighting is architecture-agnostic, easy to implement, and complementary to inference-time length control methods. It provides a practical means to fine-tune pre-trained models to generate summaries that meet specific length requirements, a crucial consideration for many real-world applications. This work thus offers a valuable contribution to the field of controllable text generation by providing an accessible tool for more precise management of output length.

\section{Limitations}

In general, the proposed method seems to effectively control generation length without compromising the quality of the generated text. However, it seems the effectiveness of the method depends on the characteristics of the underlying dataset. This is exemplified by the results on the XL-sum, whereby the heavily right-skewed distribution of summary length seems to reduce the efficacy of our method.

Secondly, the fine-tune process required for pre-trained language models incurs significant computational costs, potentially limiting the scalability and accessibility of our method compared to approaches that solely rely on inference-time operations.


\bibliography{ecai-template/main}

\appendix

\section{Complete results}\label{sec:full}
We report the complete results for the Fixed Length approach in Table~\ref{tab:results_cnn_fixed_len_complete} and Table~\ref{tab:results_xlsum_fixed_len_complete}. The complete results for the Dynamic Length approach are available in Table~\ref{tab:results_cnn_dyn_len_complete} and Table~\ref{tab:results_xlsum_dyn_len_complete} instead.

\begin{table*}
\small
\begin{center}
    \resizebox{0.95\textwidth}{!}{%
        \begin{tabular}{|l|c|c|c|c|c|c|c|c|c|c|c|c|c|c|c|}
            \hline
                & \multicolumn{2}{|c|}{Rouge-1} & \multicolumn{2}{|c|}{Rouge-2} & \multicolumn{2}{|c|}{Rouge-L} & \multicolumn{2}{|c|}{Rouge-Lsum} & \multicolumn{2}{|c|}{BertScore} & \multicolumn{2}{|c|}{\% of too long} & \multicolumn{2}{|c|}{avg. extra char} \\
                & $w$=$1$ & $w$=$10$ & $w$=$1$ & $w$=$10$ & $w$=$1$ & $w$=$10$ & $w$=$1$ & $w$=$10$ & $w$=$1$ & $w$=$10$ & $w$=$1$ & $w$=$10$ & $w$=$1$ & $w$=$10$\\
            \hline
            \multicolumn{15}{|c|}{\textbf{T5-base}} \\
            \hline
            Greedy & \textbf{34.2} & 33.2 & \textbf{14.7} & 14.5 & \textbf{26.0} & 25.7 & \textbf{30.3} & 29.6 & \textbf{26.1} & 26.0 & 9.80 & \textbf{5.4} & 2.8 & \textbf{1.2} \\
            $\text{Beam}_{-1}$ & \textbf{34.5} & 32.4 & \textbf{15.7} & 15.2 & \textbf{26.5} & 25.9 & \textbf{30.8} & 29.5 & \textbf{27.4} & 26.9 & 7.0 & \textbf{2.8} & 2.2 & \textbf{0.7} \\
            $\text{Beam}_{0}$ & \textbf{34.4} & 33.0 & 15.6 & \textbf{15.7} & 26.5 & \textbf{26.3} & 30.7 & \textbf{30.0} & 27.4 & \textbf{26.9} & 10.2 & \textbf{4.2} & 3.5 & \textbf{1.2} \\
            $\text{Beam}_{+1}$ & \textbf{35.2} & 35.0 & 15.4 & \textbf{15.7} & 25.9 & \textbf{26.5} & 30.5 & \textbf{30.9} & 25.5 & \textbf{26.1} & 57.4 & \textbf{37.0} & 43.0 & \textbf{20.4} \\
            \hline
            \multicolumn{15}{|c|}{\textbf{Llama-2-7B}} \\
            \hline
            Greedy & \textbf{38.4} & 36.5 & \textbf{17.3} & 17.0 & \textbf{28.3} & 27.9 & \textbf{36.1} & 34.3 & \textbf{34.6} & 33.2 & 7.6 & \textbf{1.8} & 1.7 & \textbf{0.5} \\
            $\text{Beam}_{-1}$ & \textbf{33.9} & 30.8 & \textbf{16.4} & 15.6 & \textbf{26.4} & 24.9 & \textbf{31.4} & 28.6 & \textbf{30.6} & 28.6 & 1.4 & \textbf{0.0} & 0.4 & \textbf{0.0} \\
            $\text{Beam}_{0}$ & \textbf{34.2} & 30.6 & \textbf{16.7} & 15.3 & \textbf{26.5} & 24.6 & \textbf{31.6} & 28.4 & \textbf{30.8} & 28.3 & 1.0 & \textbf{0.0} & 0.4 & \textbf{0.0} \\
            $\text{Beam}_{+1}$ & \textbf{34.3} & 30.9 & \textbf{16.7} & 15.5 & \textbf{26.5} & 24.8 & \textbf{31.7} & 28.7 & \textbf{30.8} & 28.6 & 2.0 & \textbf{0.0} & 1.1 & \textbf{0.0} \\
            \hline
        \end{tabular}
        }
    \vspace{10pt}
    \caption{Results for CNN/Dailymail, Fixed Length approach (250 characters).}
    \label{tab:results_cnn_fixed_len_complete}
\end{center}
\end{table*}

\begin{table*}
\small
\begin{center}
    \resizebox{0.95\textwidth}{!}{%
        \begin{tabular}{|l|c|c|c|c|c|c|c|c|c|c|c|c|c|c|c|}
            \hline
                & \multicolumn{2}{|c|}{Rouge-1} & \multicolumn{2}{|c|}{Rouge-2} & \multicolumn{2}{|c|}{Rouge-L} & \multicolumn{2}{|c|}{Rouge-Lsum} & \multicolumn{2}{|c|}{BertScore} & \multicolumn{2}{|c|}{\% of too long} & \multicolumn{2}{|c|}{avg. extra char} \\
                & $w$=$1$ & $w$=$10$ & $w$=$1$ & $w$=$10$ & $w$=$1$ & $w$=$10$ & $w$=$1$ & $w$=$10$ & $w$=$1$ & $w$=$10$ & $w$=$1$ & $w$=$10$ & $w$=$1$ & $w$=$10$\\
            \hline
            \multicolumn{15}{|c|}{\textbf{T5-base}} \\
            \hline
            Greedy & 37.0 & \textbf{38.1} & 15.6 & \textbf{16.3} & 26.5 & \textbf{27.3} & 32.1 & \textbf{33.2} & 25.5 & \textbf{26.8} & 37.2 & \textbf{18.6} & 42.5 & \textbf{7.7} \\
             $\text{Beam}_{-1}$ & 37.6 & \textbf{37.8} & 16.6 & 16.6 & 27.3 & 27.3 & 32.8 & \textbf{33.1} & 27.8 & \textbf{27.9} & 34.0 & \textbf{13.4} & 29.4 & \textbf{4.0} \\
             $\text{Beam}_{0}$ & 37.8 & \textbf{38.2} & 16.6 & \textbf{16.8} & 27.2 & \textbf{27.3} & 32.9 & \textbf{33.4} & 27.8 & \textbf{27.9} & 36.4 & \textbf{20.0} & 34.4 & \textbf{8.0} \\
             $\text{Beam}_{+1}$& 36.3 & \textbf{39.0} & 15.9 & \textbf{16.9} & 25.2 & \textbf{27.1} & 30.9 & \textbf{33.6} & 24.6 & \textbf{27.4} & 83.4 & \textbf{61.2} & 217.2 & \textbf{48.7} \\
            \hline
            \multicolumn{15}{|c|}{\textbf{Llama-2-7B}} \\
            \hline
            Greedy & \textbf{41.9} & 41.7 & \textbf{18.8} & 18.7 & \textbf{29.4} & 29.0 & \textbf{39.4} & 39.1 & \textbf{35.9} & 35.5 & 21.0 & \textbf{12.4} & 10.2 & \textbf{4.2} \\
             $\text{Beam}_{-1}$& \textbf{40.7} & 39.7 & \textbf{18.6} & 18.1 & \textbf{28.7} & 28.0 & \textbf{37.8} & 36.8 & \textbf{34.1} & 32.8 & 14.8 & \textbf{8.4} & 6.7 & \textbf{2.4} \\
             $\text{Beam}_{0}$ & \textbf{40.8} & 39.8 & \textbf{18.7} & 18.1 & \textbf{28.7} & 28.1 & \textbf{37.9} & 36.8 & \textbf{34.1} & 32.9 & 17.0 & \textbf{9.6} & 8.6 & \textbf{3.0} \\
             $\text{Beam}_{+1}$ & \textbf{40.9} & 39.8 & \textbf{18.7} & 18.10 & \textbf{28.8} & 28.0 & \textbf{38.0} & 36.9 & \textbf{34.2} & 32.80 & 19.6 & \textbf{11.0} & 9.6 & \textbf{3.5} \\
            \hline
        \end{tabular}
        }
    \vspace{10pt}
    \caption{Results for CNN/Dailymail, Dynamic Length approach.}
    \label{tab:results_cnn_dyn_len_complete}
\end{center}
\end{table*}

\begin{table*}
\small
\begin{center}
    \resizebox{0.95\textwidth}{!}{%
        \begin{tabular}{|l|c|c|c|c|c|c|c|c|c|c|c|c|c|c|c|}
            \hline
                & \multicolumn{2}{|c|}{Rouge-1} & \multicolumn{2}{|c|}{Rouge-2} & \multicolumn{2}{|c|}{Rouge-L} & \multicolumn{2}{|c|}{Rouge-Lsum} & \multicolumn{2}{|c|}{BertScore} & \multicolumn{2}{|c|}{\% of too long} & \multicolumn{2}{|c|}{avg. extra char} \\
                & $w$=$1$ & $w$=$10$ & $w$=$1$ & $w$=$10$ & $w$=$1$ & $w$=$10$ & $w$=$1$ & $w$=$10$ & $w$=$1$ & $w$=$10$ & $w$=$1$ & $w$=$10$ & $w$=$1$ & $w$=$10$\\
            \hline
            \multicolumn{15}{|c|}{\textbf{T5-base}} \\
            \hline
            Greedy & 31.2 & \textbf{31.3} & 11.9 & \textbf{12.0} & \textbf{25.4} & 25.2 & \textbf{25.4} & 25.2 & 33.2 & \textbf{33.9} & 2.0 & \textbf{0.6} & 5.9 & \textbf{1.6} \\
            $\text{Beam}_{-1}$ & 31.7 & \textbf{32.1} & 12.7 & \textbf{12.8} & 25.7 & \textbf{25.9} & 25.7 & \textbf{25.8} & 36.2 & 36.2 & 0.0 & 0.0 & 0.0 & 0.0 \\
            $\text{Beam}_{0}$ & 31.9 & \textbf{32.1} & \textbf{13.0} & 12.9 & 25.9 & 25.9 & 25.8 & \textbf{25.9} & \textbf{36.3} & 36.2 & 0.0 & 0.0 & 0.0 & 0.0 \\
            $\text{Beam}_{+1}$ & 32.0 & \textbf{32.4} & 12.9 & \textbf{13.0} & 25.7 & \textbf{25.9} & 25.7 & \textbf{25.9} & 33.1 & \textbf{34.4} & 8.2 & \textbf{4.8} & 26.0 & \textbf{10.6} \\
            \hline
            \multicolumn{15}{|c|}{\textbf{Llama-2-7B}} \\
            \hline
            Greedy & \textbf{37.9} & 37.5 & \textbf{17.5} & 16.9 & \textbf{30.6} & 30.0 & \textbf{30.6} & 30.0 & \textbf{43.9} & 43.6 & 1.0 & \textbf{0.8} & 0.8 & \textbf{0.7} \\
            $\text{Beam}_{-1}$ & \textbf{37.9} & 37.2 & \textbf{18.3} & 18.2 & \textbf{31.0} & 30.7 & \textbf{31.0} & 30.7 & 44.0 & 44.0 & 0.0 & 0.0 & 0.0 & 0.0 \\
            $\text{Beam}_{0}$ & \textbf{37.9} & 37.2 & 18.2 & 18.2 & \textbf{31.0} & 30.7 & \textbf{31.0} & 30.7 & \textbf{44.0} & 43.9 & 0.0 & 0.0 & 0.0 & 0.0 \\
            $\text{Beam}_{+1}$ & \textbf{38.0} & 37.2 & 18.3 & 18.3 & \textbf{31.0} & 30.8 & \textbf{31.0} & 30.7 & \textbf{44.0} & 43.9 & \textbf{0.0} & 0.2 & 0.0 & 0.0 \\
            \hline
        \end{tabular}
        }
    \vspace{10pt}
    \caption{Results for XLsum-multi-sentence, Fixed Length approach (175 characters).}
    \label{tab:results_xlsum_fixed_len_complete}
\end{center}
\end{table*}

\begin{table*}
\small
\begin{center}
    \resizebox{0.95\textwidth}{!}{%
        \begin{tabular}{|l|c|c|c|c|c|c|c|c|c|c|c|c|c|c|c|}
            \hline
                & \multicolumn{2}{|c|}{Rouge-1} & \multicolumn{2}{|c|}{Rouge-2} & \multicolumn{2}{|c|}{Rouge-L} & \multicolumn{2}{|c|}{Rouge-Lsum} & \multicolumn{2}{|c|}{BertScore} & \multicolumn{2}{|c|}{\% of too long} & \multicolumn{2}{|c|}{avg. extra char} \\
                & $w$=$1$ & $w$=$10$ & $w$=$1$ & $w$=$10$ & $w$=$1$ & $w$=$10$ & $w$=$1$ & $w$=$10$ & $w$=$1$ & $w$=$10$ & $w$=$1$ & $w$=$10$ & $w$=$1$ & $w$=$10$\\
            \hline
            \multicolumn{15}{|c|}{\textbf{T5-base}} \\
            \hline
            Greedy & 31.0 & \textbf{31.3} & 10.9 & \textbf{11.2} & 24.6 & \textbf{24.8} & 24.5 & \textbf{24.8} & 32.4 & \textbf{32.6} & \textbf{10.4} & 11.4 & 13.2 & \textbf{11.1} \\
            $\text{Beam}_{-1}$ & \textbf{31.2} & 31.1 & \textbf{12.5} & 12.2 & \textbf{25.2} & 25.0 & \textbf{25.1} & 24.9 & \textbf{34.4} & 34.1 & \textbf{7.2} & 7.6 & \textbf{1.5} & 2.3 \\
            $\text{Beam}_{0}$ & \textbf{31.5} & 31.3 & \textbf{12.7} & 12.3 & \textbf{25.4} & 25.1 & \textbf{25.3} & 25.0 & \textbf{34.4} & 34.2 & \textbf{7.2} & 8.0 & \textbf{1.5} & 2.4 \\
            $\text{Beam}_{+1}$ & \textbf{31.8} & 31.5 & \textbf{12.6} & 12.3 & \textbf{25.2} & 24.9 & \textbf{25.1} & 24.8 & 31.6 & \textbf{31.7} & 24.4 & \textbf{23.8} & 41.4 & \textbf{27.8} \\
            \hline
            \multicolumn{15}{|c|}{\textbf{Llama-2-7B}} \\
            \hline
            Greedy & \textbf{37.3} & 36.7 & \textbf{16.5} & 15.7 & \textbf{29.5} & 28.8 & \textbf{29.4} & 28.7 & 41.6 & 41.6 & \textbf{7.2} & 9.2 & 0.7 & 0.7 \\
            $\text{Beam}_{-1}$ & 37.4 & \textbf{37.7} & 17.1 & \textbf{17.5} & 29.9 & \textbf{30.0} & 29.8 & \textbf{29.9} & 42.0 & \textbf{42.2} & \textbf{3.4} & 3.6 & \textbf{0.2} & 0.3 \\
            $\text{Beam}_{0}$ & 37.4 & \textbf{37.6} & 17.1 & \textbf{17.5} & 29.9 & \textbf{30.0} & 29.8 & \textbf{29.9} & 41.8 & \textbf{42.2} & \textbf{3.4} & 3.8 & \textbf{0.2} & 0.3 \\
            $\text{Beam}_{+1}$ & 37.4 & \textbf{37.6} & 17.0 & \textbf{17.5} & 29.8 & \textbf{30.0} & 29.8 & \textbf{29.9} & 41.8 & \textbf{42.2} & \textbf{3.6} & 4.0 & \textbf{0.2} & 0.3 \\
            \hline
        \end{tabular}
        }
    \vspace{10pt}
    \caption{Results for XLsum-multi-sentence, Dynamic Length approach.}
    \label{tab:results_xlsum_dyn_len_complete}
\end{center}
\end{table*}

\end{document}